\documentclass[a4paper]{article}

\usepackage{INTERSPEECH2019}
\usepackage{enumitem}
\usepackage{graphics}
\usepackage{siunitx}
\usepackage{url}
\usepackage{dcolumn}
\usepackage{hyperref}

\title{Synchronising audio and ultrasound by learning cross-modal embeddings}

\name{Aciel Eshky, Manuel Sam Ribeiro, Korin Richmond, Steve Renals}

\address{
CSTR, School of Informatics, University of Edinburgh, UK
}
\email{\{aeshky, sam.ribeiro, korin.richmond, s.renals\}@ed.ac.uk}

\begin{document}

\maketitle

\begin{abstract}
Audiovisual synchronisation is the task of determining the time offset between speech audio 
and a video recording of the articulators. 
In child speech therapy, audio and ultrasound videos of the tongue are captured using instruments
which rely on hardware to synchronise the two modalities at recording time.
Hardware synchronisation can fail in practice, and no mechanism  
exists to synchronise the signals post hoc. 
To address this problem, 
we employ a two-stream neural network 
which exploits the correlation between the two modalities 
to find the offset. 
We train our model on recordings from 69 speakers, 
and show that it correctly synchronises 82.9\% of test utterances from 
unseen therapy sessions and unseen speakers,
thus considerably reducing the number of utterances to be manually synchronised. 
An analysis of model performance on the test utterances shows that 
directed phone articulations are more difficult to automatically synchronise 
compared to utterances containing natural variation in speech such as
words, sentences, or conversations. 
\end{abstract}


\section{Introduction}
Ultrasound tongue imaging (UTI) is a non-invasive way 
of observing the vocal tract during speech production \cite{stone2005guide}. 
Instrumental speech therapy relies on 
capturing ultrasound videos of the patient's tongue
simultaneously with their speech audio in order to
provide a diagnosis, design treatments, 
and measure therapy progress \cite{cleland2015using}.
The two modalities must be correctly synchronised,
with a minimum shift of $+$45ms if the audio leads and $-$125ms if the audio lags,
based on synchronisation standards for broadcast audiovisual signals~\cite{bt1359relative}. 
Errors beyond this range can render the data unusable -- indeed, synchronisation errors do occur, 
resulting in significant wasted effort if not corrected.
No mechanism currently exists to automatically correct these errors, and 
although manual synchronisation is possible 
in the presence of certain audiovisual cues such as stop consonants \cite{alm2013audio},
it is time consuming and tedious. 

In this work, we exploit the correlation between the two modalities to synchronise them.
We utilise a two-stream neural network architecture for the task \cite{chung2016out}, 
using as our only source of supervision pairs of ultrasound and audio segments 
which have been automatically generated and labelled as
positive (correctly synchronised) or negative (randomly desynchronised); 
a process known as self-supervision \cite{korbar2018cooperative}.
We demonstrate how this approach enables us to correctly synchronise the majority of 
utterances in our test set,
and in particular, those exhibiting natural variation in speech.

\begin{figure}[t]
  \centering
  \includegraphics[height=0.65\columnwidth]{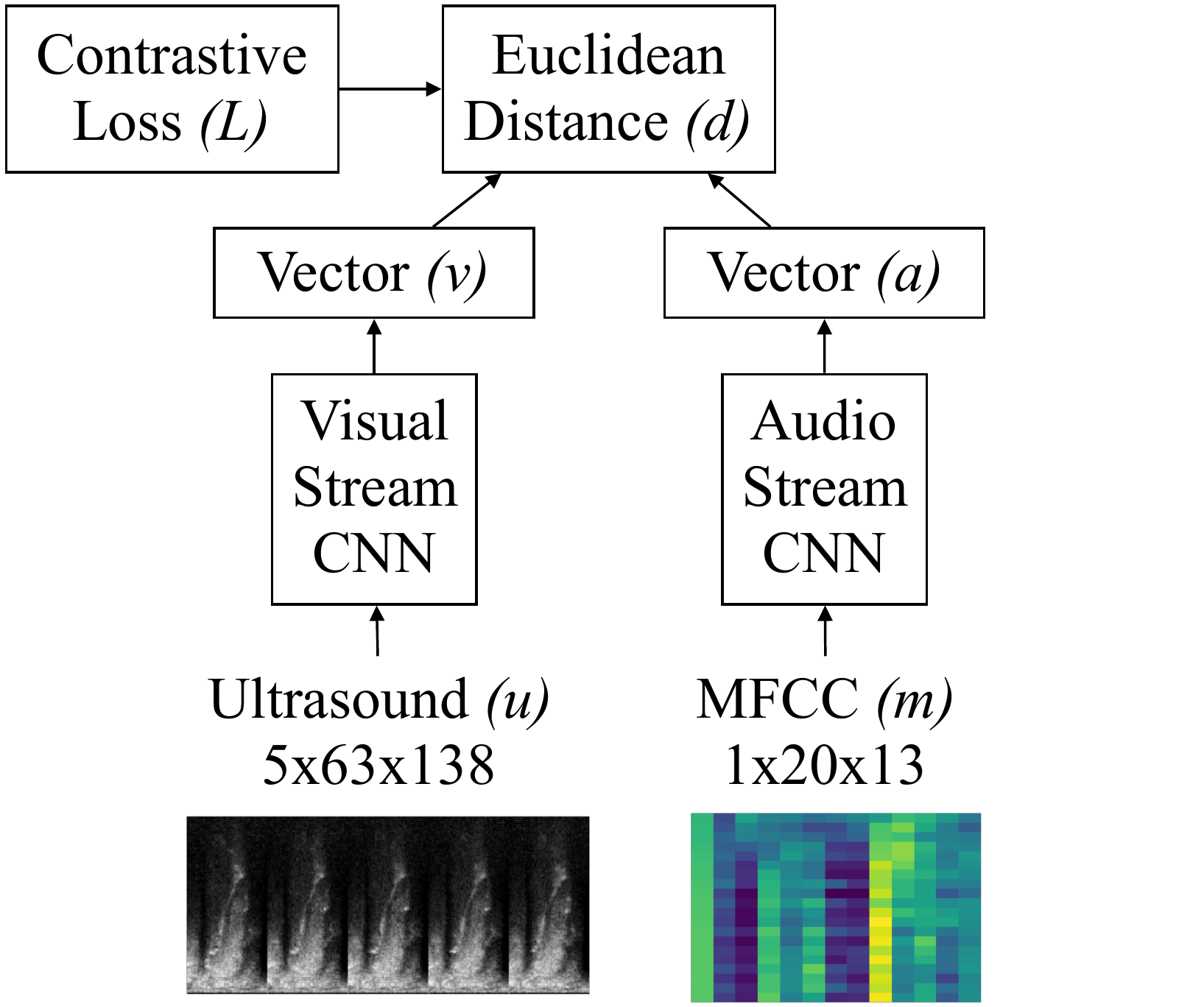}
  \caption{UltraSync maps high dimensional inputs to low dimensional vectors using a contrastive loss function, 
  such that the Euclidean distance is small between vectors from positive pairs and large otherwise. 
  Inputs span $\simeq$200ms: 5 consecutive raw ultrasound frames on one stream and 20 frames of the corresponding MFCC features on the other.}
  \label{fig:model_diagram}
\end{figure}

Section~\ref{sec:background} reviews existing approaches for audiovisual synchronisation, 
and describes the challenges specifically associated with UTI data, compared with lip videos
for which automatic synchronisation has been previously attempted. 
Section~\ref{sec:model} describes our approach.
Section~\ref{sec:data} describes the data we use, 
including data preprocessing 
and positive and negative sample creation using a self-supervision strategy.
Section~\ref{sec:experiments} describes our experiments,
followed by an analysis of the results. 
We conclude with a summary and future directions in Section~\ref{sec:conclusion}\footnote{Code available at: {\url{https://github.com/aeshky/ultrasync}}}.

\section{Background}\label{sec:background}
Ultrasound and audio are recorded using separate components, 
and hardware synchronisation is achieved by translating information from the visual 
signal into audio at recording time.
Specifically, for every ultrasound frame recorded, 
the ultrasound beam-forming unit releases a pulse signal,
which is translated by an external hardware synchroniser 
into an audio pulse signal and captured by the sound card \cite{wrench2018sono, wrench2018articulate}. 
Synchronisation is achieved by aligning the ultrasound frames with the audio pulse signal, 
which is already time-aligned with the speech audio \cite{wrench2018personal}. 

Hardware synchronisation can fail for a number of reasons.
The synchroniser is an external device which needs to be correctly connected and operated 
by therapists. Incorrect use can lead to missing the pulse signal, 
which would cause synchronisation to fail 
for entire therapy sessions \cite{cleland2018personal}. 
Furthermore, low-quality sound cards report an approximate, 
rather than the exact, sample rate which leads to errors 
in the offset calculation \cite{wrench2018personal}.
There is currently no recovery mechanism for when synchronisation fails,
and to the best of our knowledge, there has been no prior work on automatically 
correcting the synchronisation error between ultrasound tongue videos and audio. 
There is, however, some prior work on synchronising lip movement with audio 
which we describe next.

\subsection{Audiovisual synchronisation for lip videos}
Speech audio is generated by articulatory movement
and is therefore fundamentally correlated with other
manifestations of this movement, such as lip or tongue videos
\cite{yehia1998quantitative}. 
An alternative to the hardware approach 
is to exploit this correlation to find the offset.
Previous approaches have investigated the effects of using different representations 
and feature extraction techniques on finding dimensions of high correlation
\cite{sargin2007audiovisual, bredin2007audiovisual, garau2010audio}.
More recently, neural networks, which learn features directly from input, 
have been employed for the task.
SyncNet \cite{chung2016out} 
uses a two-stream neural network and self-supervision 
to learn cross-modal embeddings, 
which are then used to synchronise audio with lip videos. 
It achieves near perfect accuracy ($>$99$\%$) using manual evaluation 
where lip-sync error is not detectable to a human. 
It has since been extended 
to use different sample creation methods for self-supervision \cite{korbar2018cooperative, chung2019perfect}
and different training objectives \cite{chung2019perfect}.
We adopt the original approach \cite{chung2016out}, 
as it is both simpler and significantly less expensive 
to train than the more recent variants.

\subsection{Lip videos vs. ultrasound tongue imaging (UTI)} \label{sec:limitations}
Videos of lip movement can be obtained 
from various sources including TV, films, and YouTube, 
and are often cropped to include only the lips \cite{chung2016out}. 
UTI data, on the other hand, is recorded in clinics by trained therapists \cite{eshky2018ultrasuite}.
An ultrasound probe placed under the chin of the patient 
captures the midsaggital view of their oral cavity as they speak. 
UTI data consists of sequences of 2D matrices of raw ultrasound reflection data,
which can be 
interpreted as greyscale images \cite{eshky2018ultrasuite}.
There are several challenges specifically associated with UTI data 
compared with lip videos,
which can potentially lower the performance of models relative to 
results reported on lip video data. These include:

\textbf{Poor image quality:} 
Ultrasound data is noisy, containing arbitrary high-contrast edges, speckle noise, artefacts,
and interruptions to the tongue's surface \cite{stone2005guide, li2005automatic, ribeiro2019speaker}.  
The oral cavity is not entirely visible, missing the lips, the palate, 
and the pharyngeal wall, and visually interpreting the data requires specialised training.
In contrast, videos of lip movement are of much higher quality
and suffer from none of these issues.

\textbf{Probe placement variation:} 
Surfaces that are orthogonal to the ultrasound beam image 
better than those at an angle. Small shifts in probe placement 
during recording lead to high variation between otherwise 
similar tongue shapes \cite{stone2005guide, cleland2018ultrax2020, ribeiro2019speaker}. 
In contrast, while the scaling and rotations of lip videos 
lead to variation, they do not lead to a degradation in image quality.

\textbf{Inter-speaker variation:} 
Age and physiology affect the quality of ultrasound data, and
subjects with smaller vocal tracts and less tissue fat image better 
\cite{stone2005guide, ribeiro2019speaker}.
Dryness in the mouth, as a result of nervousness during speech therapy, 
leads to poor imaging. 
While inter-speaker variation is expected in lip videos, again, the variation
does not lead to quality degradation. 

\textbf{Limited amount of data:} 
Existing UTI datasets are considerably smaller than lip movement datasets.
Consider for example VoxCeleb and VoxCeleb2 used to train SyncNet \cite{chung2016out, chung2019perfect},
which together contain 1 million utterances from 7,363 identities
\cite{Nagrani17, Chung18b}. 
In contrast, the UltraSuite repository (used in this work) contains 
13,815 spoken utterances from 86 identities.

\textbf{Uncorrelated segments:} 
Speech therapy data contains interactions between the therapist and patient. 
The audio therefore contains speech from both speakers, 
while the ultrasound captures only the patient's tongue \cite{eshky2018ultrasuite}. 
As a result, parts of the recordings will consist of completely uncorrelated
audio and ultrasound. This issue is similar to that of dubbed voices in lip videos \cite{chung2016out},
but is more prevalent in speech therapy data.

\section{Model}\label{sec:model}
We adopt the approach in \cite{chung2016out}, modifying it to synchronise audio with UTI data. 
Our model, UltraSync, consists of two streams: 
the first takes as input a short segment of ultrasound
and the second takes as input the corresponding audio.
Both inputs are high-dimensional and are of different sizes. 
The objective is to learn a mapping from the inputs 
to a pair of low-dimensional vectors of the same length, 
such that the Euclidean distance  
between the two vectors is small when they correlate and large otherwise \cite{chopra2005learning, hadsell2006dimensionality}. 
This model can be viewed as an extension of a siamese neural network
\cite{bromley1994signature} 
but with two asymmetrical streams and no shared parameters. 
Figure~\ref{fig:model_diagram} illustrates the main architecture.
The visual data $u$ (ultrasound) and audio data $m$ (MFCC),
which have different shapes, are mapped 
to low dimensional embeddings $v$ (visual) and $a$ (audio)
of the same size:
\begin{equation}
\psi(u; \theta) \rightarrow v,\ 
\phi(m; \eta) \rightarrow a
\end{equation}

The model is trained using 
a contrastive loss function \cite{chopra2005learning, hadsell2006dimensionality}, 
$L$, which minimises the Euclidean distance 
$d = || v - a ||_2$ between $v$ and $a$ for positive pairs ($y=1$), 
and maximises it for negative pairs ($y=0$), for a number of training samples $N$: 
\begin{equation}\label{equation_E}
L(\theta, \eta) = \frac{1}{N} \sum^{N}_{n=1}
y_n d_n^2 + (1-y_n)\{max(1-d_n, 0)\}^2
\end{equation}

Given a pair of ultrasound and audio segments 
we can calculate the distance between them using our model. 
To predict the synchronisation offset for an utterance, 
we consider a discretised set of candidate offsets,
calculate the average distance for each across utterance segments, 
and select the one with the minimum average distance. 
The candidate set is independent of the model, and is chosen 
based on task knowledge (Section~\ref{sec:experiments}).

\section{Data}\label{sec:data}
For our experiments, we select a dataset 
whose utterances have been correctly synchronised at recording time.
This allows us to control how the model is trained and 
verify its performance using ground truth synchronisation offsets.
We use UltraSuite\footnote{\url{http://www.ultrax-speech.org/ultrasuite}}:
a repository of ultrasound and acoustic data gathered from child speech therapy sessions \cite{eshky2018ultrasuite}.
We used all three datasets from the repository:
UXTD (recorded with typically developing children), and UXSSD and UPX (recorded with children with speech sound disorders).
In total, the dataset contains 13,815 spoken utterances from 86 speakers, corresponding to 35.9 hours of recordings.
The utterances have been categorised by the type of task the child was given, and are labelled as:
Words (A), Non-words (B), Sentence (C), 
Articulatory (D), Non-speech (E), or Conversations (F). See \cite{eshky2018ultrasuite} for details.

Each utterance consists of 3 files: audio, ultrasound, and parameter.
\textbf{The audio file} is a RIFF wave file, sampled at 22.05 KHz, containing the speech of the child and therapist. 
\textbf{The ultrasound file} consists of a sequence of ultrasound frames 
capturing the midsagittal view of the child's tongue.  
A single ultrasound frame is recorded as a 2D matrix where
each column represents the ultrasound reflection intensities along a single scan line.
Each ultrasound frame consists of 63 scan lines of 412 data points each,
and is sampled at a rate of $\simeq$121.5 fps. 
Raw ultrasound frames can be visualised as greyscale 
images and can thus be interpreted as videos. 
\textbf{The parameter file} contains the synchronisation offset value (in milliseconds),
determined using hardware synchronisation at recording time and 
confirmed by the therapists to be correct for this dataset. 

\subsection{Preparing the data}
First, we exclude utterances of type ``Non-speech" (E) from our training data (and statistics). 
These are coughs recorded to obtain additional tongue shapes, 
or swallowing motions recorded to capture a trace of the 
hard palate. Both of these rarely contain audible content and are therefore not relevant to our task.
%
Next, we apply the offset, which should be positive if the audio leads and negative if the audio lags.
In this dataset, the offset is always positive.
We apply it by cropping the leading audio
and trimming the end of the longer signal to match the duration. 

To process the ultrasound more efficiently, 
we first reduce the frame rate from $\simeq$121.5 fps to $\simeq$24.3 fps by retaining 1 out of every 5 frames.
%
We then downsample by a factor of (1, 3), shrinking the frame size from 63x412 to 63x138 using max pixel value. 
This retains the number of ultrasound vectors (63), but reduces the number of pixels per vector (from 412 to 138). 

The final pre-preprocessing step is to remove empty regions. 
UltraSuite was previously anonymised by zero-ing segments 
of audio which contained personally identifiable information. 
As a preprocessing step, we remove the zero regions from audio 
and corresponding ultrasound. 
We additionally experimented with removing regions of silence using voice activity detection, 
but obtained a higher performance by retaining them.

\subsection{Creating samples using a self-supervision strategy}

To train our model we need positive and negative training pairs. 
The model ingests short clips from each modality of $\simeq$200ms long, 
calculated as $t = l/r$, where 
$t$ is the time window,
$l$ is the number of ultrasound frames per window (5 in our case), 
and $r$ is the ultrasound frame rate of the utterance ($\simeq$24.3 fps). 
For each recording, we split the ultrasound into non-overlapping windows of 5 frames each.
We extract MFCC features (13 cepstral coefficients) from the audio
using a window length of $\simeq$20ms, calculated as $t/(l\times2)$,
and a step size of $\simeq$10ms, calculated as $t/(l\times4)$. 
This give us the input sizes shown in Figure~\ref{fig:model_diagram}.

Positive samples are pairs of ultrasound windows and the corresponding MFCC frames. 
To create negative samples, we randomise pairings of 
ultrasound windows to MFCC frames within the same utterance,
generating as many negative as positive samples to achieve a balanced dataset. 
We obtain 
243,764 samples for UXTD (13.5hrs), 
333,526 for UXSSD (18.5hrs), and
572,078 for UPX (31.8 hrs),
or a total 1,149,368 samples (63.9hrs)
which we divide into training, validation and test sets. 

\subsection{Dividing samples for training, validation and testing}
We aim to test whether our model generalises 
to data from new speakers,
and to data from new sessions recorded with known speakers.
To simulate this, we select a group of speakers from each dataset, 
and hold out all of their data either for validation or for testing. 
Additionally, we hold out one entire session from each of the remaining speakers,
and use the rest of their data for training. 
We aim to reserve approximately 80\% of the created samples for training, 
10\% for validation, and 10\% for testing,
and select speakers and sessions on this basis. 

Each speaker in UXTD recorded 1 session, but sessions are of different durations. 
We reserve 45 speakers for training, 5 for validation, and 8 for testing.
UXSSD and UPX contain fewer speakers, but each recorded multiple sessions.
We hold out 1 speaker for validation and 1 for testing from each of the two datasets. 
We also hold out a session from the first half of the remaining speakers for validation, 
and a session from the second half of the remaining speakers for testing\footnote{Held out subsets: 
UXTD speakers 07, 08, 12, 13, 26 for validation 
and speakers 30, 38, 43, 45, 47, 52, 53, 55 for testing.
UXSSD speaker 01 and session `Mid' (for speakers 02-04) for validation, 
and speaker 07 and session `Mid' (for speakers 05, 06, 08) for testing.
UPX speakers 01 and session `BL3' (for speakers 02-10) for validation, 
and speaker 15 and session `BL3' (for speakers 11-14 and 16-20) for testing.}.
This selection process results in 909,858 (pooled) samples for training (50.5hrs), 
128,414 for validation (7.1hrs) and 111,096 for testing (6.2hrs). 
From the training set, we create shuffled batches which are 
balanced in the number of positive and negative samples.

\section{Experiments}\label{sec:experiments}

\begin{table}[t]
\caption{
Each stream has 3 convolutional layers followed by 2 fully-connected layers.
Fully connected layers have 64 units each.
For convolutional layers, we specify the number of filters and their receptive 
field size as ``num$\times$size$\times$size" 
followed by the max-pooling downsampling factor.  
Each layer is followed by batch-normalisation then ReLU activation. 
Max-pooling is applied after the activation function.}
\label{lab:parameters_table}
\centering
\begin{tabular}{lrrrrr}
\toprule
\textbf{Stream} & \textbf{Conv1} & \textbf{Conv2} & \textbf{Conv3} & \textbf{Full4} & \textbf{Full5}\tabularnewline
\midrule
Visual & 23x5x5 & 64x5x5 & 128x5x5 & 64 & 64\tabularnewline
 & x2 pool & x2 pool & x2 pool &  & \tabularnewline
\midrule
Audio & 23x3x3 & 64x3x3 & 128x3x3 & 64 & 64\tabularnewline
 &  & x2 pool & x2 pool &  & \tabularnewline
\bottomrule
\end{tabular}
\end{table}
We select the hyper-parameters of our model empirically by tuning on the validation set
(Table~\ref{lab:parameters_table}). 
Hyper-parameter exploration is guided by \cite{chatfield2014return}.
We train our model using the Adam optimiser \cite{kingma2015adam} with a learning rate of 0.001,
a batch size of 64 samples, and for 20 epochs. 
We implement learning rate scheduling which reduces
the learning rate by a factor of 0.1 when the validation loss plateaus for 2 epochs.

%


\begin{table*}[t]
\caption{
Model accuracy per test set and utterance type. 
Performance is consistent across test sets for Words (A) where the sample sizes are large,
and less consistent for types where the sample sizes are small. 
71\% of UXTD utterances are Articulatory (D), which explains the low performance on this test set
(64.8\% in Table~\ref{lab:results_1}).
In contrast, performance on UXTD Words (A) is comparable to other test sets.
}
\label{lab:results_3}
\centering
\begin{tabular}{lrrrrrrrrrr}
\toprule
\textbf{} & 
\multicolumn{2}{c}{\textbf{Words (A)}} & 
\multicolumn{2}{c}{\textbf{Non-words (B)}} &
\multicolumn{2}{c}{\textbf{Sentence (C)}} & 
\multicolumn{2}{c}{\textbf{Articulatory (D)}} & 
\multicolumn{2}{c}{\textbf{Conversation (F)}}\\
\textbf{Test Set} & 
\multicolumn{1}{r}{\textbf{N}} & \multicolumn{1}{r}{\textbf{Acc}} & 
\multicolumn{1}{r}{\textbf{N}} & \multicolumn{1}{r}{\textbf{Acc}} & 
\multicolumn{1}{r}{\textbf{N}} & \multicolumn{1}{r}{\textbf{Acc}} &
\multicolumn{1}{r}{\textbf{N}} & \multicolumn{1}{r}{\textbf{Acc}} & 
\multicolumn{1}{r}{\textbf{N}} & \multicolumn{1}{r}{\textbf{Acc}}\\
\midrule
UXTD&108&88.9\%&22&86.4\%&0&-&325&55.4\%&0&-\\
UXSSD&307&88.6\%&20&65.0\%&58&94.8\%&11&54.5\%&0&-\\
UPX&499&92.4\%&16&100.0\%&128&93.8\%&4&100.0\%&4&75.0\%\\
\midrule
All&914&90.7\%&58&82.8\%&186&94.1\%&340&55.9\%&4&75.0\%\\
\bottomrule
\end{tabular}
\end{table*}

\begin{table}[t]
\caption{Model accuracy per utterance type, where N is the number of utterances.
Performance is best on utterances containing natural variation in speech,
such as Words (A) and Sentences (C). 
Non-words (B) and Conversations (F) also exhibit this variation, 
but due to smaller sample sizes the lower percentages are not representative. 
Performance is lowest on Articulatory utterances (D), which contain isolated phones.
The mean and standard deviation of the discrepancy 
between the prediction and the true offset are also shown in milliseconds.}
\label{lab:results_2} 
\centering
\begin{tabular}{lrrr@{}l}
\toprule
\multicolumn{1}{l}{\textbf{Utterance Type}} & 
\multicolumn{1}{r}{\textbf{N}} & 
\multicolumn{1}{r}{\textbf{Acc}} & 
\multicolumn{2}{r}{\textbf{Discrepancy}}\\
\midrule 
Words (A)			&914		&90.7\%&		1 $\pm$ 		&~102 ms\\
Non-words (B)		&58		&82.8\%&		$-$2 $\pm$ 	&~~~39 ms\\
Sentence (C)		&186		&94.1\%&		16 $\pm$ 		&~150 ms\\
Articulatory (D)		&340		&55.9\%&		129 $\pm$ 	&~408 ms\\
Conversation (F)	&4		&75.0\%&		$-$87 $\pm$ 	&~141 ms\\
\midrule
All				&1502	&82.9\%&		32 $\pm$ 		&~223 ms\\
A, B, C and F		&1162	&91.2\%&		3 $\pm$ 		&~112 ms\\
\bottomrule 
\end{tabular}
\end{table}

\begin{table}[t]
\caption{
Model accuracy per test set.
Contrary to expectation, 
performance is better on test sets containing new speakers than on
test sets containing new sessions from known speakers.
The performance on UXTD is considerably lower than other test sets, 
due to it containing a large number of Articulatory utterances, 
which are difficult to synchronise 
(see Tables~\ref{lab:results_2}~and~\ref{lab:results_3}).}
\label{lab:results_1}
\centering
\begin{tabular}{lrrr@{}l}
\toprule
\multicolumn{1}{l}{\textbf{Test Set}} & 
\multicolumn{1}{r}{\textbf{N}} & 
\multicolumn{1}{r}{\textbf{Acc}} & 
\multicolumn{2}{r}{\textbf{Discrepancy}}\\
\midrule 
UXTD,  new speakers	&455		&64.8\%	& 97 $\pm$ 	&~357 ms\\
UXSSD, new sessions	&126		&82.5\%	& 19 $\pm$ 	&~160 ms\\
UXSSD, new speaker	&270		&89.6\%	& 9 $\pm$ 	&~135 ms\\
UPX, new sessions		&306		&91.2\%	& $-$3 $\pm$	&~~~40 ms\\
UPX, new speaker		&345		&94.2\%	& $-$2 $\pm$ 	&~123 ms\\
\midrule
All					&1502	&82.9\%	& 32 $\pm$ 	&~223 ms\\
\bottomrule 
\end{tabular}
\end{table}

Upon convergence, the model achieves 
0.181 training loss, 
0.213 validation loss, and 
0.213 test loss. 
By placing a threshold of 0.5 on predicted distances, the model achieves 
72.7\% binary classification accuracy on training samples, 
65.6\% on validation samples, and
65.6\% on test samples.

\textbf{Synchronisation offset prediction:} 
Section~\ref{sec:model} described briefly how to use our model
to predict the synchronisation offset for test utterances. 
To obtain a discretised set of offset candidates, 
we retrieve the true offsets of the training utterances,
and find that they fall in the range [0, 1789] ms. 
We discretise this range taking 45ms steps 
and rendering 40 candidate values 
(45ms is the smaller of the absolute values 
of the detectability thresholds, $-$125 and $+$45 ms~\cite{bt1359relative}).
We bin the true offsets in the candidate set 
and discard empty bins, reducing the set from 40 to 24 values.
We consider all 24 candidates for each test utterance. 
We do this by aligning the two signals according to the given candidate,
then producing the non-overlapping windows of ultrasound and MFCC pairs, 
as we did when preparing the data. 
We then use our model to predict the Euclidean distance for each pair, and average the distances.
Finally, we select the offset with the smallest average distance as our prediction.

\textbf{Evaluation:} 
Because the true offsets are known, we evaluate the performance of our model by computing the 
discrepancy between the predicted and the true offset for each utterance. 
If the discrepancy falls within the detectability thresholds~\cite{bt1359relative}
($-$125 $<$ $x$ $<$ $+$45) then the prediction is correct. 
Random prediction (averaged over 1000 runs) yields 14.6\% accuracy 
with a mean and standard deviation discrepancy of 328 $\pm$ 518ms.
We achieve 82.9\% accuracy
with a mean and standard deviation discrepancy of 32 $\pm$ 223ms.
SyncNet reports $>$99\% accuracy on lip video synchronisation 
using a manual evaluation where the lip error is not detectable to a human observer \cite{chung2016out}.
However, we argue that our data is more challenging (Section~\ref{sec:limitations}).

\textbf{Analysis:}
We analyse the performance of our model across different conditions.
Table~\ref{lab:results_2} shows the model accuracy broken down by utterance type.
The model achieves 91.2\% accuracy on utterances 
containing words, sentences, and conversations, 
all of which exhibit natural variation in speech.
The model is less successful with Articulatory utterances,
which contain isolated phones occurring once or repeated (e.g., ``sh sh sh").
Such utterances contain subtle tongue movement,
making it more challenging to correlate the visual signal with the audio.
And indeed, the model finds the correct offset for only 55.9\% of Articulatory utterances.
A further analysis shows that 84.4\% (N$=$90) 
of stop consonants (e.g., ``t''), which are relied upon by therapists 
as the most salient audiovisual synchronisation cues \cite{alm2013audio},
are correctly synchronised by our model,
compared to 48.6\% (N$=$140) of vowels, which contain less distinct movement
and are also more challenging for therapists to synchronise. 

Table~\ref{lab:results_1} shows accuracy broken down by test set.
The model performs better on test sets containing entirely new speakers
compared with test sets containing new sessions from previously seen speakers.
This is contrary to expectation but could be due
to the UTI challenges (described in Section~\ref{sec:limitations})
affecting different subsets to different degrees.
Table~\ref{lab:results_1} shows that the model performs considerably worse 
on UXTD compared to other test sets (64.8\% accuracy). 
However, a further breakdown of the results in Table~\ref{lab:results_3} 
by test set and utterance type explains this poor performance;
the majority of UXTD utterances (71\%) are Articulatory utterances
which the model struggles to correctly synchronise. 
In fact, for other utterance types (where there is a large enough sample, such as Words)
performance on UXTD is on par with other test sets. 

\section{Conclusion}\label{sec:conclusion}
We have shown how a two-stream neural network 
originally designed to synchronise lip videos with audio 
can be used to synchronise UTI data with audio.
Our model exploits the correlation between the modalities to
learn cross-model embeddings which are used to find the synchronisation offset.
It generalises well to held-out data,
allowing us to correctly synchronise the majority of test utterances.
The model is best-suited to utterances which contain natural variation in speech
and least suited to those containing isolated phones, with the exception of stop consonants.  
Future directions include integrating the model and synchronisation offset prediction process
into speech therapy software \cite{wrench2018sono, wrench2018articulate},
and using the learned embeddings for other tasks 
such as active speaker detection \cite{chung2016out}.

\section{Acknowledgements} 
{ 
Supported by EPSRC Healthcare Partnerships Programme grant number 
EP/P02338X/1 (Ultrax2020). } 

\bibliographystyle{IEEEtran}

\bibliography{sync_paper_biblio}

\end{document}